%% file: main.tex
\documentclass[conference]{IEEEtran}
\IEEEoverridecommandlockouts
\usepackage{cite}
\usepackage{amsmath,amssymb,amsfonts}
\usepackage{algorithmic}
\usepackage{graphicx}
\usepackage{textcomp}
\usepackage{kotex}
\usepackage{graphicx}
\usepackage{subcaption}
\usepackage{float}
\usepackage{booktabs}
\usepackage{hyperref}

\def\BibTeX{{\rm B\kern-.05em{\sc i\kern-.025em b}\kern-.08em
    T\kern-.1667em\lower.7ex\hbox{E}\kern-.125emX}}

\title{Urban Vibrancy Embedding and Application on Traffic Prediction}

\usepackage{kotex}
\usepackage{graphicx}
\usepackage{subcaption}
\usepackage{float}
\usepackage{booktabs}
\begin{document}

%%
%% The "title" command has an optional parameter,
%% allowing the author to define a "short title" to be used in page headers.
% \title{Embedding Urban Vibrancy Knowledge and Adaption for Enhanced Traffic Prediction: A Case Study}
\title{Urban Vibrancy Embedding and Application on Traffic Prediction}

% \title{Enhancing %Urban Dynamics 
% \textcolor{red}{Traffic
% Prediction 
% Performance
% }
% by Embedding Urban Vibrancy Knowledge %-- Traffic Prediction Case Study
% }

%%
%% The "author" command and its associated commands are used to define
%% the authors and their affiliations. 
%% Of note is the shared affiliation of the first two authors, and the
%% "authornote" and "authornotemark" commands
%% used to denote shared contribution to the research.

\author{
    \begin{minipage}[t]{0.3\textwidth}
        \centering
        {Sumin Han}\\
        \textit{School of Computing}\\
        KAIST\\
        Daejeon 34141, South Korea\\
        suminkaist@gmail.com\\
        ORCID: 0000-0002-4071-8469
    \end{minipage}
    \hfill
    \begin{minipage}[t]{0.3\textwidth}
        \centering
        {Jisun An}\\
        \textit{Luddy School of Informatics}\\
        Indiana Univ. Bloomington\\
        Bloomington, Indiana, USA\\
        jisunan@iu.edu\\
        ORCID: 0000-0002-4353-8009
    \end{minipage} 
    \begin{minipage}[t]{0.3\textwidth}
        \centering
        {Dongman Lee}\\
        \textit{School of Computing}\\
        KAIST\\
        Daejeon 34141, South Korea\\
        dlee@kaist.ac.kr\\
        ORCID: 0000-0001-5923-6227
    \end{minipage}
}

\maketitle

\input{sec0-abstract}

%%
%% The code below is generated by the tool at http://dl.acm.org/ccs.cfm.
%% Please copy and paste the code instead of the example below.


\begin{figure*}[t]
  \centering
  \includegraphics[width=\textwidth]{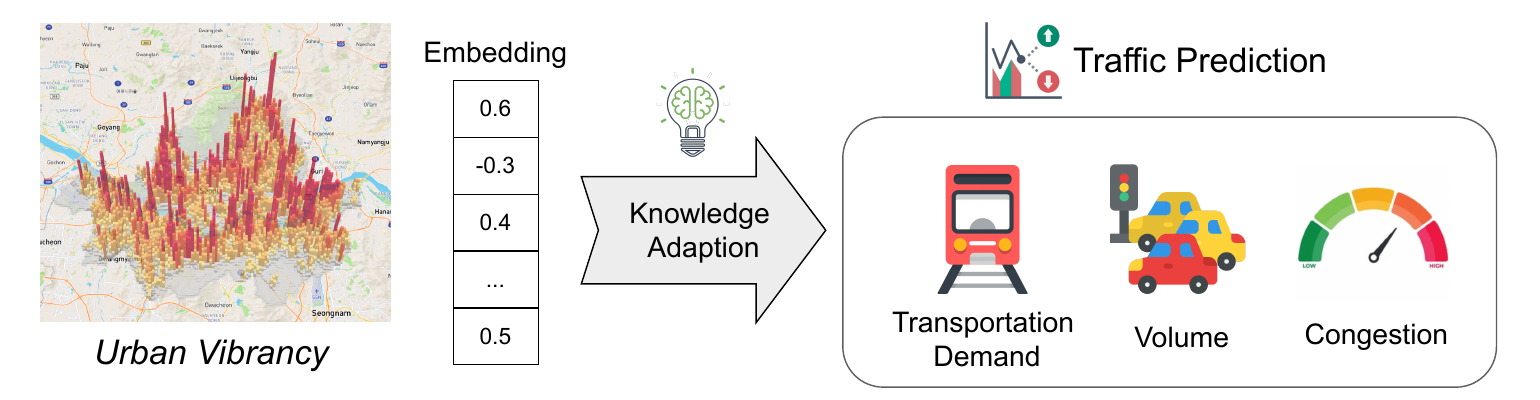}
  \caption{Embedding Urban Vibrancy and Knowledge Adaption on Traffic Prediction.}
  \label{fig:teaser}
\end{figure*}

% \received{20 February 2007}
% \received[revised]{12 March 2009}
% \received[accepted]{5 June 2009}

%%
%% This command processes the author and affiliation and title
%% information and builds the first part of the formatted document.
\maketitle

\input{sec1-intro}

\input{sec2-related}
\input{sec3-pre}

\input{sec4-method}
\input{sec5-experiment}

\input{sec6-results}

\input{sec7-conclusion}

%%
%% The acknowledgments section is defined using the "acks" environment
%% (and NOT an unnumbered section). This ensures the proper
%% identification of the section in the article metadata, and the
%% consistent spelling of the heading.
\section*{Acknowledgement}
This work was supported by the Institute of Information \& Communications Technology Planning \& Evaluation (IITP) grant funded by the Korea government (MSIT) (No. RS-2019-II191126, Self-learning based Autonomic IoT Edge Computing) and National Research Foundation (NRF) funded by the Korean government (MSIT) (No.RS-2024-00356597).

%%
%% The next two lines define the bibliography style to be used, and
%% the bibliography file.
\bibliographystyle{ieeetr}
\bibliography{
    bibfile.bib
}

%%
%% If your work has an appendix, this is the place to put it.
% \appendix

% \input{sec8-appendix}

\end{document}

%% file: sec0-abstract.tex
\begin{abstract}

Urban vibrancy reflects the dynamic human activity within urban spaces and is often measured using mobile data that captures floating population trends. This study proposes a novel approach to derive Urban Vibrancy embeddings from real-time floating population data to enhance traffic prediction models. Specifically, we utilize variational autoencoders (VAE) to compress this data into actionable embeddings, which are then integrated with long short-term memory (LSTM) networks to predict future embeddings. These are subsequently applied in a sequence-to-sequence framework for traffic forecasting. Our contributions are threefold: (1) We use principal component analysis (PCA) to interpret the embeddings, revealing temporal patterns such as weekday versus weekend distinctions and seasonal patterns; (2) We propose a method that combines VAE and LSTM, enabling forecasting dynamic urban knowledge embedding; and (3) Our approach improves accuracy and responsiveness in traffic prediction models, including RNN, DCRNN, GTS, and GMAN. This study demonstrates the potential of Urban Vibrancy embeddings to advance traffic prediction and offer a more nuanced analysis of urban mobility.

\end{abstract}

%% file: sec1-intro.tex
\section{Introduction}

Urban vibrancy reflects the dynamic human activities that occur within urban spaces, and it is increasingly measured through mobile data capturing floating population\footnote{The floating population refers to individuals who temporarily reside in a specific area without being counted in the official census, often for work or educational purposes.} trends\cite{bergroth202224,lee2018utilizing}. This concept embodies not just the visible energy of a city but also the potential for interactions and the fundamental social conditions that shape urban life \cite{barreca2020urban}. In terms of human activity, urban vibrancy can be expressed as the intensity and diversity of people's interactions within a city, providing a lens into how individuals engage with their environment on both physical and social levels \cite{fu2021relationship}.

Understanding urban vibrancy poses significant challenges, primarily in terms of \textbf{Interpretation} and \textbf{Application}.

1. \textbf{Interpretation}: A major challenge lies in effectively clustering and interpreting patterns of urban vibrancy as observed through floating population data, requiring a systematic approach to categorize patterns and establish criteria for similarity. Conventional methods, such as spatiotemporal analysis \cite{lee2018urban, lee2018utilizing}, kernel density estimation \cite{zhang2021understanding}, and point-of-interest (POI) association \cite{gao2024exploring}, have been developed to capture and characterize urban vibrancy. However, these approaches often struggle with the demands of real-time data streams and lack the granularity needed to interpret complex, high-dimensional, or image-based datasets.

2. \textbf{Application}: Another challenge is determining how urban vibrancy derived from floating population data can be applied in various domains. Recent works have explored applications in urban vibrancy similar to multi-task learning. For example, research on citywide crowd flow prediction has used spatial-temporal neural networks to model long-range dependencies \cite{feng2021context}, while other studies have employed deep multi-view networks for tasks like taxi demand prediction \cite{yao2018deep}. These applications underscore the potential of urban vibrancy data, yet they have not fully harnessed the insights that real-time floating population trends can offer for traffic prediction and similar dynamic applications.

This study introduces an innovative method for deriving Urban Vibrancy embeddings from real-time floating population data to enhance traffic prediction models. Specifically, we employ variational autoencoders (VAE) to compress this data into actionable embeddings. These embeddings are then integrated with long short-term memory (LSTM) networks to predict future states and applied within a sequence-to-sequence framework for traffic forecasting.

Our contributions are as follows:

\begin{enumerate}
    \item Interpretation of Real-Time Embeddings: Through detailed principal component analysis (PCA), we visualize the generated embeddings, which reveal distinct temporal patterns such as weekday versus weekend trends, hourly variations, and seasonal shifts. This analysis provides intuitive insights into the underlying dynamics of urban vibrancy.
    
    \item Dynamic Urban Vibrancy Embedding: By integrating VAEs and LSTMs, our approach embeds real-time urban data dynamically, enhancing the model's adaptability to changing urban conditions and improving its responsiveness to real-time data fluctuations.
    
    \item Enhanced Model Responsiveness and Accuracy: Our method demonstrates substantial improvements in traffic prediction models, including RNN, DCRNN, GTS, and GMAN, by incorporating real-time urban insights. This makes it particularly suitable for the dynamic demands of smart city environments.
    
    \item Comprehensive Experimental Validation: We validate our approach on real-world data from Seoul, Korea, showcasing the method's effectiveness. We also provide our code and dataset to the community to support reproducibility and further research \footnote{\href{https://github.com/suminhan/wka-net}{https://github.com/suminhan/wka-net}}.
\end{enumerate}

This paper contributes to the field by offering a novel perspective on using urban vibrancy to drive advanced traffic prediction, facilitating a more nuanced and responsive approach to urban mobility analysis.

%% file: sec2-related.tex
\section{Related Work}

\subsection{Urban Vibrancy using Floating Population}

Urban vibrancy derived from floating population has become an increasingly important aspect of urban studies and planning. This concept refers to the dynamic energy and vitality that temporary or non-permanent residents bring to urban areas\cite{barreca2020urban}. Floating populations, which include tourists, business travelers, and short-term migrants, contribute significantly to the diversity and intensity of human activities in cities\cite{chen2022investigating}. These transient groups often engage in various social, economic, and cultural activities, thereby enhancing the overall vibrancy of urban spaces\cite{tu2020portraying}. Researchers have utilized multisource urban big data, including mobile phone location data and social media check-ins, to analyze the spatiotemporal patterns of floating populations and their impact on urban vibrancy\cite{lee2024does, jang2024urban}. Such studies have revealed that floating populations can significantly influence the attractiveness of neighborhoods, the use of public spaces, and the overall economic vitality of urban areas6. Understanding the relationship between floating populations and urban vibrancy is crucial for urban planners and policymakers, as it can inform decisions related to infrastructure development, public service provision, and strategies for enhancing the overall quality of urban life\cite{barreca2020urban}.

% Urban Vibrancy에 대한 연구를 적는다.
% In the domain of traffic prediction, conventional origin-destination-based travel forecasting models have faced criticism for their limited understanding of nuanced travel behaviors~\cite{rasouli2012uncertainty, schneider2013unravelling}. In response, activity-based models, rooted in the inseparable link between travel and human activities~\cite{ben1998activity, rasouli2014activity, pinjari2011activity, pas1988weekly, recker1985travel}, have gained prominence. These models help to consider factors like individual travel destinations, schedules, and purposes, recognizing that urban residents engage in daily activities at varying times and locations~\cite{hanson1981travel, neutens2011prism, bhat2004comprehensive}. To comprehensively grasp population-wide travel behaviors, an understanding of the urban context, including population distribution and employment locations, is crucial~\cite{hanson2004context, schonfelder2016urban}. 
% %This knowledge enables the explicit representation of spatiotemporal urban travel patterns, contributing to more effective traffic prediction~\cite{castiglione2015activity}. 
% The incorporation of elements from activity-based approaches, such as 24-hour temporal travel patterns, route preferences, and activity-space constraints~\cite{hasnine2021tour, doherty200212}, can further enhances prediction accuracy, bridging the gap between human activity and transportation modeling.

\subsection{Knoweldge Adaption on Traffic Prediction}

Conventional traffic prediction models like DCRNN \cite{DCRNN}, ASTGCN \cite{ASTGCN}, and GraphWaveNet \cite{GraphWavenet} incorporate external information as an additional input channel alongside traffic data but lack mechanisms to process this information into interpretable embeddings. While \cite{knowledge-adaption-cikm} introduces a neural network for cross-mode knowledge adaptation, it does not extend to graph-based spatio-temporal models. The MultiView Deep LSTM framework, designed for ride-hailing demand forecasting, captures features from order, speed, and weather views but lacks sufficient interpretability and does not fully exploit complex graph structures in spatio-temporal data \cite{MVDLSTM}.

%% file: sec3-pre.tex
\section{Preliminaries}

The traffic prediction problem addressed in this study is defined as follows: The traffic value is represented as \( X_{t} \in \mathbb{R}^{n_s \times n_f} \), where \( n_s \) denotes the number of sensors, and \( n_f \) represents multiple channels, such as "on" and "off" in the case of demand. Let \( C_t \in \mathbb{R}^{n_w \times n_h \times n_c} \) represent the real-time floating population at timestamp \( t \), where \( n_w \) and \( n_h \) are the width and height of the area, respectively, and \( n_c \) indicates the number of channels. Consequently, the traffic prediction task is defined as a \( p \)-to-\( q \) sequence prediction finding an optimal function \( h \):

$$ h(X_{p-t+1}, ..., X_{t}, C_{p-t+1}, ..., C_{t}) \rightarrow \hat{X}_{t+1}, ..., \hat{X}_{t+q} $$

%% file: sec4-method.tex
\section{Methods}

In this study, we employ a Variational Autoencoder to extract Urban Vibrancy Embeddings, which are then used for traffic prediction. Specifically, we apply an LSTM-based sequence-to-sequence model to forecast future Urban Vibrancy Knowledge. The predicted embeddings are fed into the Decoder component of the pretrained Variational Autoencoder, where the error between the predicted and actual Floating Population values is calculated for training. Finally, we incorporate an Urban Knowledge-Informed Spatio-Temporal Embedding into existing models such as RNN, DCRNN, GTS, and GMAN, which we hypothesize will improve overall model performance.

Our Urban Vibrancy Knowledge extraction approach is inspired by the Vision-Memory-Control (VMC) framework proposed in \cite{ha2018worldmodels}, where components are divided into vision, memory, and control stages. In our method, embeddings of the current floating population scene serve as the “vision” component, and these embeddings are used to predict future scenes, replacing the control component with a prediction mechanism in a structure similar to VMC.

\subsection{Variational Autoencoder (VAE)} \label{sec:vae}

\begin{figure}[h]
    \centering
    \includegraphics[width=\columnwidth]{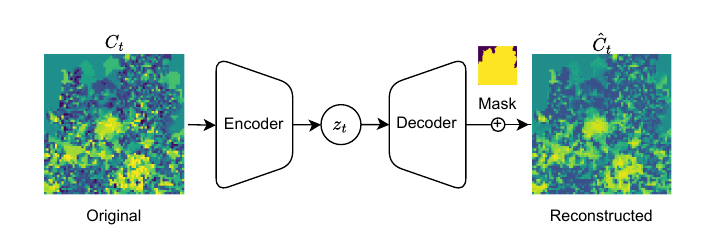}
    \caption{Variational Autoencoder}
    \label{fig:vae}
\end{figure}

The Floating Population data is normalized to values between 0 and 1, similar to image pixel values, and represented in a grid format. Details regarding the preprocessing of the $C$ values are outlined in Section~\ref{sec:floating-population-preprocessing}.

As illustrated in Figure~\ref{fig:vae}, the Variational Autoencoder (VAE) model is applied to the refined Floating Population image at time $\tau$, denoted as $C_\tau$, to extract the embedding $z_\tau$ (i.e., $z_\tau \leftarrow \text{Encoder}(C_\tau) \in \mathbb{R}^{d}$). The Decoder then reconstructs $\hat{C}_\tau$ (i.e., $\hat{C}_\tau \leftarrow \text{Decoder}(z_\tau + \textit{Noise})$), and the model is trained using Mean Squared Error (MSE) loss between $C_\tau$ and $\hat{C}_\tau$. The VAE process involves the following steps:

\[
\mathcal{L}_{\text{vae}} = \text{MSE}(C, \hat{C}) + D_{\text{KL}}(q_\phi(z|C) \parallel p(z))
\]

In detail, an image mask with values of either 0 or 1 is applied to the final layer of the decoder, where these values indicate active or inactive floating population signals. This mask is concatenated into the final layer of the Decoder, followed by an additional dense layer to produce the final output. The mask helps the model avoid misinterpreting the data distribution and prevents training on null patterns from inactive cells.

\subsection{LSTM for Forecasting Urban Vibrancy Embedding}

\begin{figure*}[t]
    \centering
    \includegraphics[width=\textwidth]{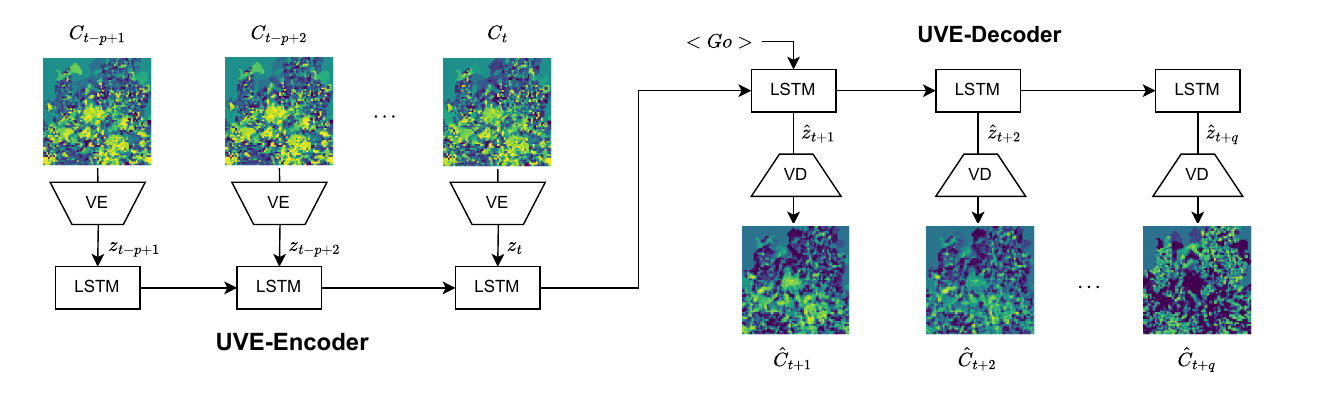}
    \caption{UVE-Seq2Seq: Forecasting Future Urban Vibrancy Embedding.} \label{fig:uve-lstm}
\end{figure*}

The UVE-Seq2Seq model, depicted in Figure~\ref{fig:uve-lstm}, leverages the VAE’s Encoder (VE) and Decoder (VD) components to predict future Urban Vibrancy Embeddings  ($\hat{z}_{t+1}, ..., \hat{z}_{t+q}$). Trained in Section~\ref{sec:vae}, the model uses embeddings derived from the current Floating Population data to forecast future embeddings. During this process, the parameters of VE and VD remain frozen and are not updated, with only the Seq2Seq LSTM model undergoing training.

To predict these embeddings, the model first extracts an embedding \( z \) for each \( C_\tau \) in the range \([t-p+1, ..., t]\):
\[
z_\tau \leftarrow \textbf{VE}(C_\tau)
\]

Using these embeddings, the UV-ENCDEC model generates future embeddings \( (\hat{z}_{t+1}, ..., \hat{z}_{t+q}) \):
\[
(\hat{z}_{t+1}, ..., \hat{z}_{t+q}) \leftarrow \textbf{UV-ENCDEC}(z_{t-p+1}, ..., z_t)
\]

Each predicted embedding \( \hat{z}_\tau \) is then transformed into an anticipated scene \( \hat{C}_\tau \) using the VAE Decoder (VD):
\[
\hat{C}_\tau \leftarrow \textbf{VD}(\hat{z}_\tau) \quad \text{for} \quad \tau \in [t+1, ..., t+q]
\]

The model optimizes its predictions using Mean Squared Error (MSE) loss:
\[
\mathcal{L}_{\text{UV-ENCDEC}} = \text{MSE}(C, \hat{C})
\]

This sequence-to-sequence LSTM model is designed to capture and predict the future distributions of Floating Population, enhancing traffic prediction by embedding urban vibrancy data into the forecasting process.

% \subsection{Profiling Knowledge Adaptability}

% \begin{figure}[h]
%     \centering
%     \includegraphics[width=\columnwidth]{images/urban-vibrancy/profiling-knowledge-adaptability-regression-process.png}
%     \caption{Profiling Knowledge Adaptability}
%     \label{fig:profiling}
% \end{figure}

% 매 Timestep마다의 Embedding값인 \( z^{(n)} \) (그림에는 X라고 잘못 나와있음) 과 실제 $i$번 센서의 교통 값인 \( y_i^{(n)}\)와의 상관관계를 Linear Regression의 R2 값을 이용해 상관관계를 구한다. 이를 통해 $N \times 168$ 의 매 시간요일에 따라 Urban Vibrancy Embedding과 센서가의 상관관계가 가장 높았던 것을 기록하는 Table이 만들어지게 된다. 이것은 다음 장에 소개할 Urban Vibrancy Knowledge Adaption을 할 때 특정한 센서에 특정한 요일 시간에 대해서만 값을 가져오도록 모델링 할 때 활용된다.

\subsection{Urban Vibrancy Knowledge Adaptation for Traffic Prediction}

While some models incorporate external information by simply concatenating it with the existing input data, we found that this approach did not lead to significant performance improvements. Therefore, we developed the Spatio-Temporal Vibrancy Embedding (STVE) as a more effective way to adapt external knowledge. This embedding is then integrated into traffic prediction models to enhance their ability to utilize urban vibrancy information.

\subsubsection{Spatio-Temporal Vibrancy Embedding (STVE)}

The extracted Urban Vibrancy Embedding (UVE) is derived from past Floating Population scenes for the previous \( p \) timesteps and forecasted for future \( q \) timesteps using the UV-ENCDEC model. This UVE is incorporated into traffic prediction models as supplementary information through Spatio-Temporal Embedding.

The concept of Spatio-Temporal Embedding, which plays a role similar to Positional Encoding in Transformers, is also utilized in works such as \cite{UAGCRN,GMAN}. It introduces markers that account for spatial and temporal differences. Here, we integrate Urban Vibrancy Embedding with temporal information to enhance traffic prediction.

The components of Spatio-Temporal Vibrancy Embedding are as follows:
\begin{itemize}
    \item \( \textbf{Z}_S \in \mathbb{R}^{n_s \times n_s} \): Sensor encoding, an identity matrix of size \( n_s \times n_s \) , which is a unique one-hot embeddings for each sensor.
    \item \( \textbf{Z}_T \in \mathbb{R}^{(p+q) \times (7+24)} \): Timestep embedding, where the day-of-week and time-of-day (24 hours) for each timestep are encoded using one-hot encoding into vectors of sizes \( \mathbb{R}^7 \) and \( \mathbb{R}^{24} \), respectively.
    \item \( \textbf{Z}_V = (z_{t-p+1}, \dots, z_t, \hat{z}_{t+1}, \dots, \hat{z}_{t+q}) \in \mathbb{R}^{(p+q) \times d} \): Urban Vibrancy Embedding, consisting of embeddings from both observed and forecasted data.
\end{itemize}

The final Spatio-Temporal Urban Vibrancy Embedding (\( \textbf{Z}_{STV} \)) combines these components as follows:
\[
\textbf{Z}_{STV} = f_1(\textbf{Z}_S) + f_2(\textbf{Z}_T || \textbf{Z}_V) \in \mathbb{R}^{n_s \times (p+q) \times d}
\]

where each $f_1$, $f_2$ is a 2-stacked fully connected layers with batch normalization, and $||$ denotes concatenation.

\subsubsection{Application to Traffic Prediction Models} \label{subsec-pretrain-traf-pred}

The Urban Knowledge-Informed Spatio-Temporal Embedding is integrated into various traffic prediction models as follows:
\begin{enumerate}
    \item \textbf{GRU}: \( \textbf{Z}_{TV} = f( \textbf{Z}_T || \textbf{Z}_V) \in \mathbb{R}^{(p+q)\times d} \) is concatenated with the input at each GRU step to enhance temporal context.
    \item \textbf{DCRNN, GTS}: In the Encoder, each input timestep \( X \) is expanded to \( d \)-dimensions using fully connected layers. These transformed inputs are then combined with urban vibrancy embedding as \( f(X) + \textbf{Z}_{STV, 1:p} \in \mathbb{R}^{n_s \times p \times d} \). In the Decoder, \( \textbf{Z}_{STV,p+1:p+q} \) is utilized in place of the GO token.
    \item \textbf{GMAN}: \( \textbf{Z}_{STV} \) replaces the traditional Spatio-Temporal Encoding (STE) to enrich the model with urban vibrancy information.
\end{enumerate}

By incorporating STVE into these models, we enhance their ability to predict traffic patterns with an additional layer of urban context, leveraging real-time vibrancy information to improve accuracy and robustness. Detailed implementations can be referred from our public repository.

%% file: sec5-experiment.tex
\begin{figure*}[t]
  \centering
  \begin{subfigure}[t]{0.25\textwidth}
    \centering
    \includegraphics[width=\textwidth]{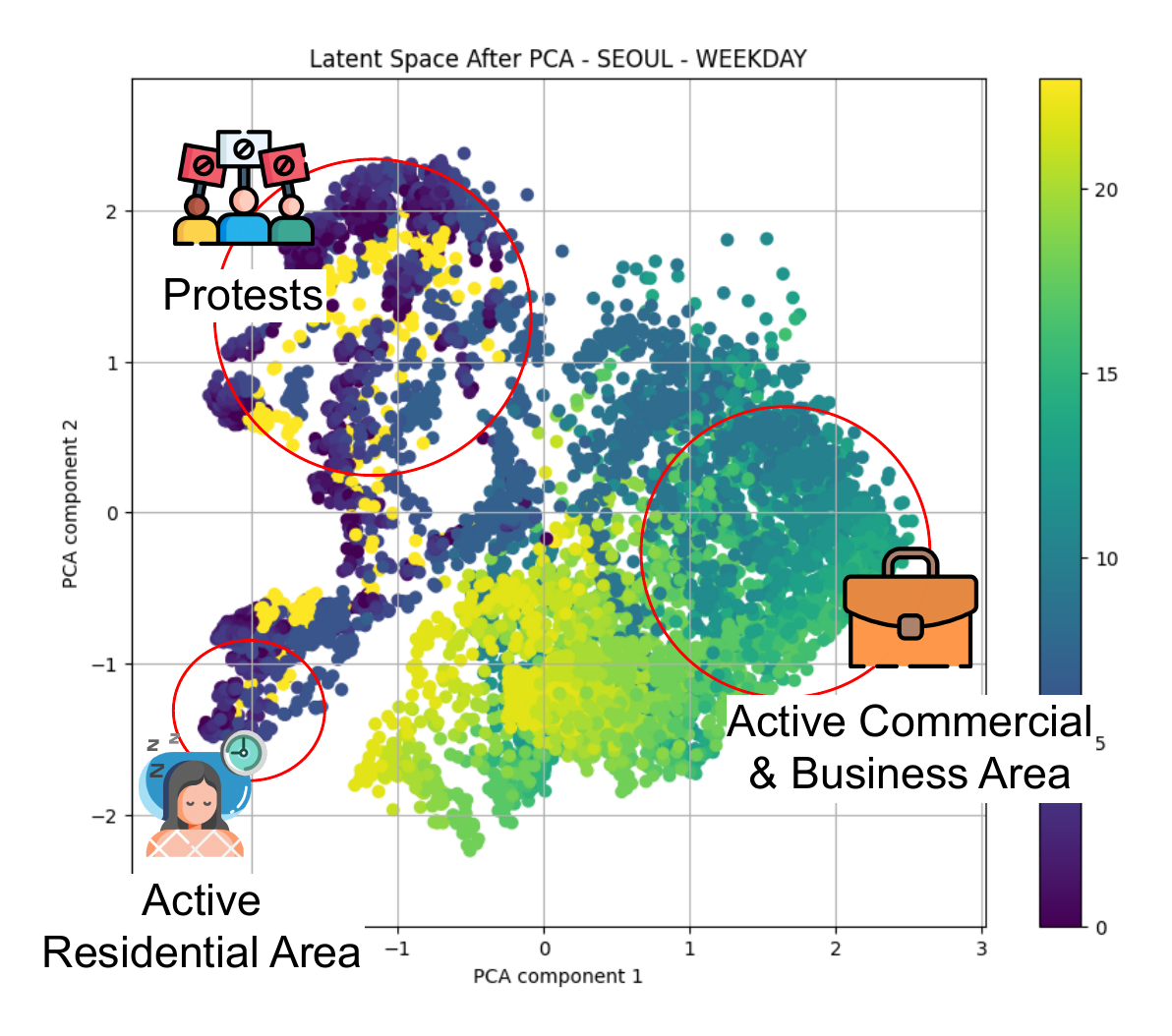} % images/pca-seoul-weekday.png
    \caption{Weekday}
  \end{subfigure}
  \hfill
  \begin{subfigure}[t]{0.25\textwidth}
    \centering
    \includegraphics[width=\textwidth]{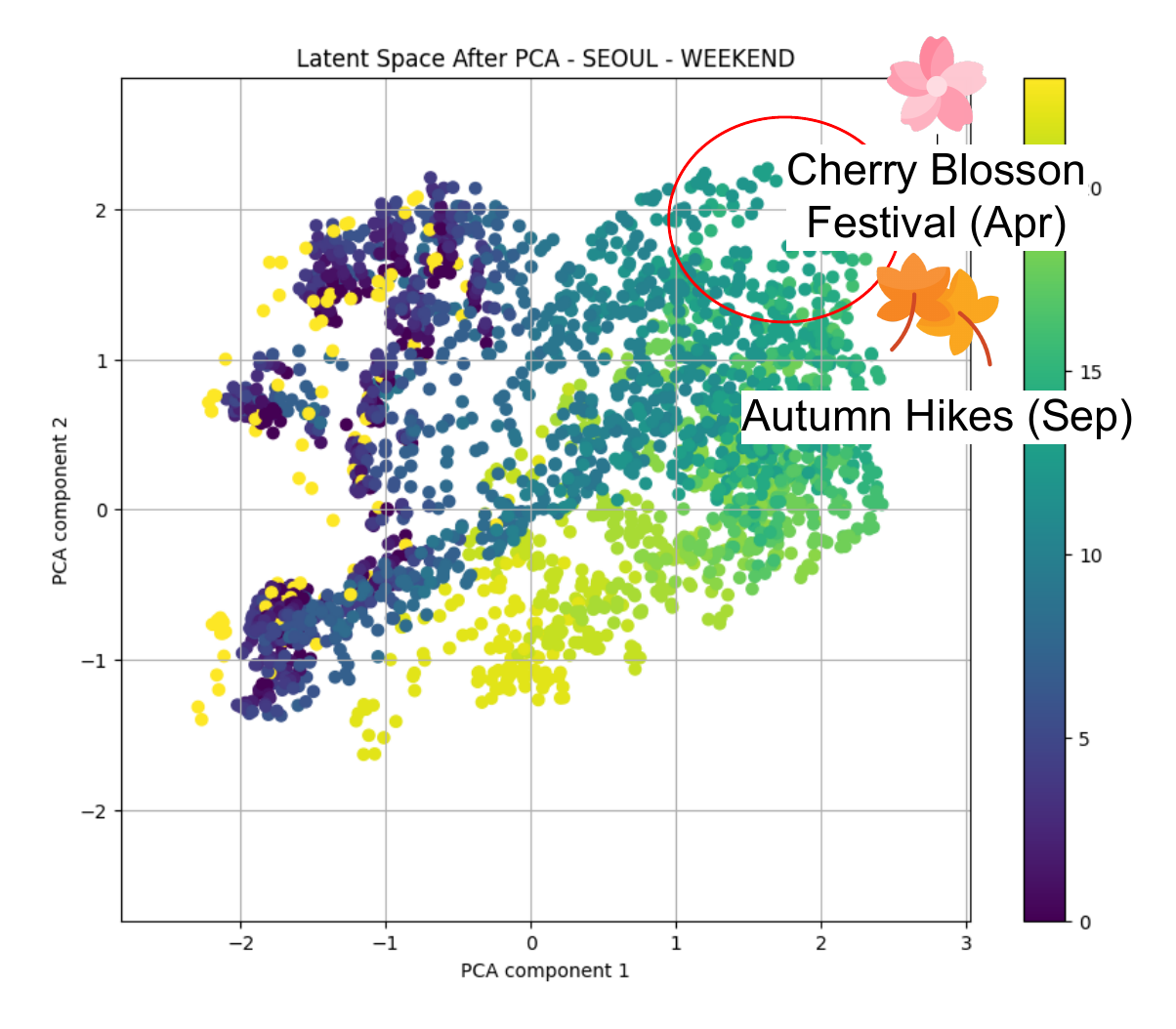} %images/pca-seoul-weekend.png
    \caption{Weekend}
  \end{subfigure}
  \hfill
  \begin{subfigure}[t]{0.24\textwidth}
    \centering
    \includegraphics[width=\textwidth]{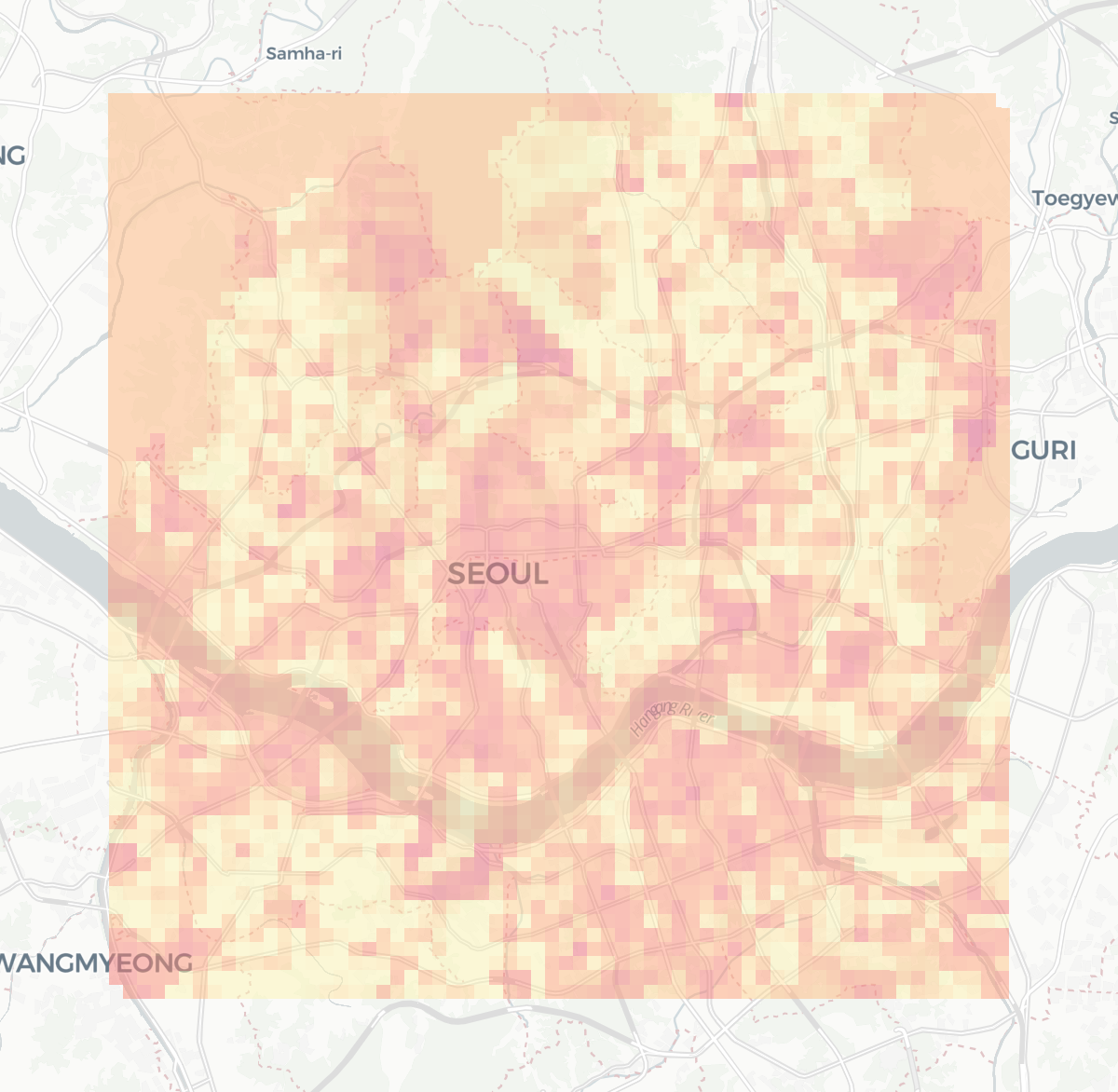}
    \caption{PCA Component 1 (x-axis): central Seoul areas.}
    \label{fig:pca-component-1}
  \end{subfigure}
  \hfill
  \begin{subfigure}[t]{0.24\textwidth}
    \centering
    \includegraphics[width=\textwidth]{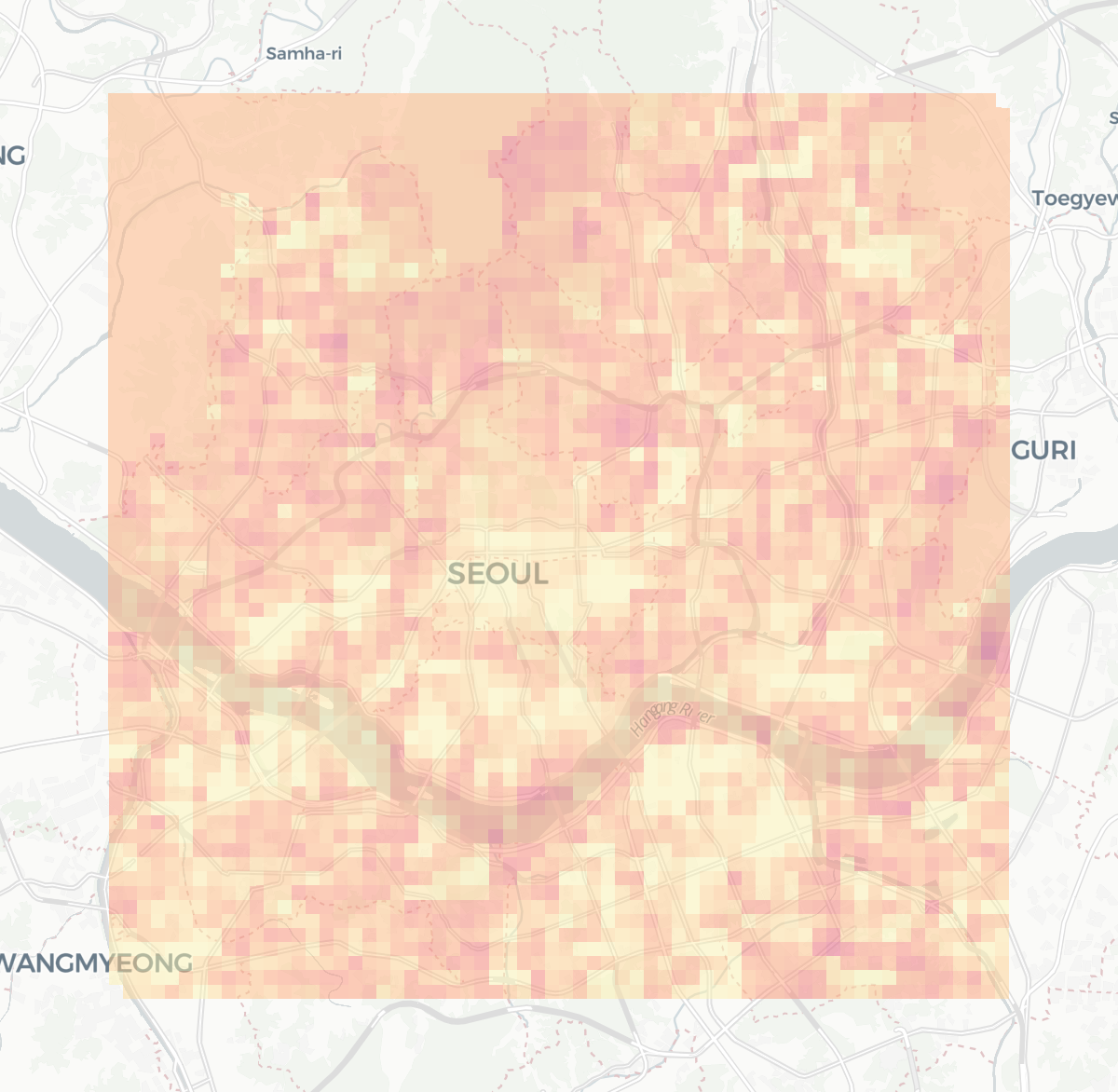}
    \caption{PCA Component 2 (y-axis): outer Seoul areas.}
    \label{fig:pca-component-2}
  \end{subfigure}
  \caption{Comparison of Urban Vibrancy Embedding on Weekdays and Weekends}
  \label{fig:urban-vibrancy-pca-component}
\end{figure*}

\section{Experimental Settings}

% \subsection{Dataset} \label{sec:floating-population-preprocessing}

% 데이터셋의 경우 서울시의 생활인구를 활용하였다. 서울시의 생활인구는 Census 단위로 집계가 된다. 
% 여기서 x1, y1, x2, y2 = [126.87214296,  37.47232319, 127.12196293,  37.67097132] 이렇게 되는 지역을 뽑아 활용한다. 이 지역은 서울시에서도 가장 중심부에 있는 가로 세로 22km 정도에 해당하는 영역이며, 강남과 강북을 포함한 서울의 중심부를 나타낸다.

% 본 연구에서는 2017년과 2018년의 각각 365일의 24시간, 총 $2 \times 365 \times 24$ 데이터를 활용하였다. 

% 교통 예측 성능향상을 검증하기 위해 지하철 승하차 데이터, 교통량 데이터, 교통속도 데이터를 각각 정제하여 활용하였다.

% Data Imputation의 경우 비어있는 값이 있을 경우 그 전주의 데이터와 그 다음주의 데이터를 평균내어 비어있는 값을 채웠다.

% 데이터에 대한 statistics는 Table~\ref{tab:datastats}에 설명되어 있다.

\subsection{Dataset} \label{sec:floating-population-preprocessing}

This study utilizes a comprehensive dataset from Seoul, which captures floating population data over a two-year period (2017-2018) across a central 22 km by 22 km area that includes major urban districts such as Gangnam and Gangbuk.\footnote{The region is defined by \([126.8721, 37.4723, 127.1219, 37.6709]\).} The land use of our research area is also depicted in Figure~\ref{fig:seoul-landuse}. The dataset records hourly variations in population density, amounting to $2 \times 365 \times 24$ observations, and is supplemented with subway ridership, traffic volume, and traffic speed data to evaluate improvements in predictive accuracy. Rigorous preprocessing was performed, with missing values imputed through temporal averaging of adjacent weeks to maintain data continuity and integrity. Table~\ref{tab:datastats} presents key statistics for each data source, highlighting variability and providing essential context for model development. This high-resolution dataset lays a robust foundation for embedding real-time urban vibrancy insights into traffic prediction, enabling an in-depth exploration of how dynamic urban activity patterns impact traffic flows in one of the world’s most vibrant metropolitan areas.

\input{tables/tab-datastats}

\subsubsection{Preprocessing Floating Population}

To preprocess the Floating Population variable \( C \), we first apply z-score normalization for each cell individually using its respective mean (\( \mu_{ij} \)) and standard deviation (\( \sigma_{ij} \)) from original values $C^o \in \mathbb{R}^{8760 \times 64 \times 64 \times 1}$: \( C'_{ij} = \frac{C^o_{ij} - \mu_{ij}}{\sigma_{ij}} \). The normalized result is then capped within ±2 standard deviations: \( C'' = \min(\max(C', -2), 2) \). Finally, the data is scaled to the [0, 1] range using \( C = \frac{C'' + 2}{4} \). % Floating Population in our research is illustrated in Figure~\ref{fig:population}.

\subsubsection{Graph Construction}

The graphs for Subway Demand, Traffic Volume, and Traffic Speed are constructed as follows: Subway Demand graphs connect adjacent stations and transfer stations, while Traffic Volume graphs link traffic sensors based on their proximity. For Traffic Speed, road segments are connected according to their physical distances.

\subsection{Settings}
We used the LibCity library\footnote{https://libcity.ai/} for our experiments. The model was trained with parameters \( p = 6 \) and \( q = 6 \), with each unit representing one hour. For each model, we applied the default learning rate and other hyperparameters provided by LibCity, ensuring that weekday and time-of-day information were consistently utilized.

The baseline models included RNN, DCRNN \cite{DCRNN}, GTS \cite{GTS}, and GMAN \cite{GMAN}. The dataset was divided into training, validation, and test sets with a ratio of 0.4, 0.1, and 0.5, respectively. The training and validation sets consisted of data from 2017, while the test set used only 2018 data. This setup ensured that the model was tested on a distinct, future year to evaluate its predictive performance.

Unlike traditional traffic prediction approaches, where data is split sequentially, we shuffled the training and validation sets within the 2017 data. This shuffling ensured that the validation set encompassed data from all seasons, rather than being biased toward winter at the end of the year. By including seasonal diversity, we aimed for a model that learns a balanced representation across various patterns and conditions throughout the year. The evaluation metric used for this experiment was Mean Absolute Error (MAE).

%% file: tables/tab-datastats.tex
\begin{table}[t]
\caption{Data Statistics} \label{tab:datastats}
\vspace{-2mm}
\begin{tabular}{@{}c|p{0.17\columnwidth}p{0.17\columnwidth}p{0.17\columnwidth}p{0.17\columnwidth}@{}}
\toprule
          & \begin{tabular}[c]{@{}c@{}}Floating\\ Population\end{tabular} & \begin{tabular}[c]{@{}c@{}}Subway\\ Demand\end{tabular} & \begin{tabular}[c]{@{}c@{}}Traffic\\ Volume\end{tabular} & \begin{tabular}[c]{@{}c@{}}Traffic\\ Speed\end{tabular} \\ \midrule
\# Nodes & $64 \times 64$                                                & 275                                                     & 73                                                       & 263                                                     \\
Channel   & 1                                                             & 2                                                       & 1                                                        & 1                                                       \\
Mean      & 570.99                                                        & 725.47                                                  & 1866.93                                                  & 28.25                                                   \\
Std       & 1092.88                                                       & 1129.04                                                 & 1480.88                                                  & 13.77                                                   \\
Timespan  & \multicolumn{4}{c}{Jan 1, 2017 $\sim$ Dec 31, 2018 (hourly, 17,520 steps)}                                                                                                                                                              \\\bottomrule
\end{tabular}
\end{table}

\begin{figure}[t]
\centering
\includegraphics[width=\columnwidth]{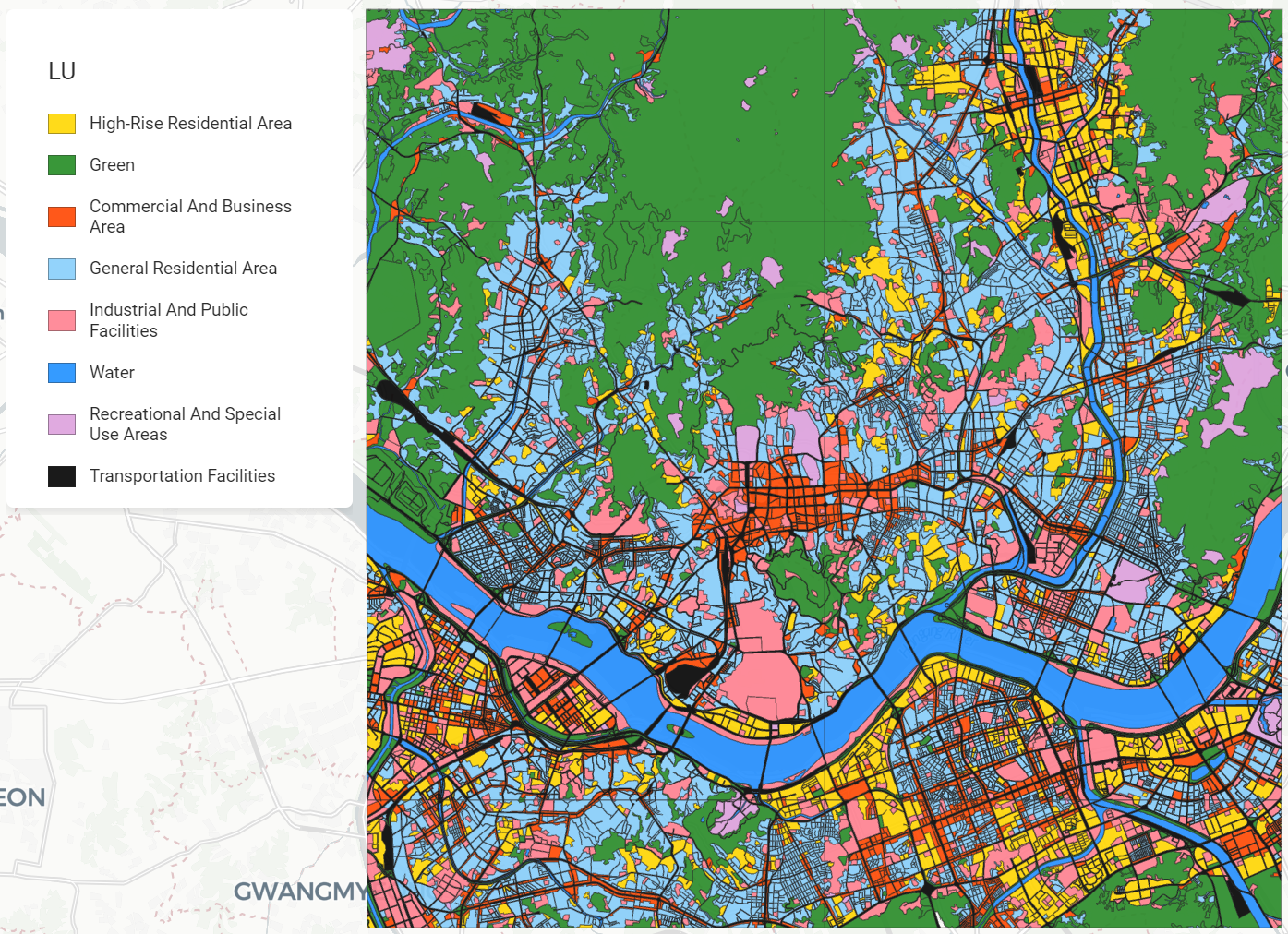}
\caption{Land Use of our Research Area}
\label{fig:seoul-landuse}
\end{figure}

%% file: sec6-results.tex
\begin{figure*}[t]
  \centering
  \begin{subfigure}[t]{0.49\textwidth}
    \centering
    \includegraphics[width=\textwidth]{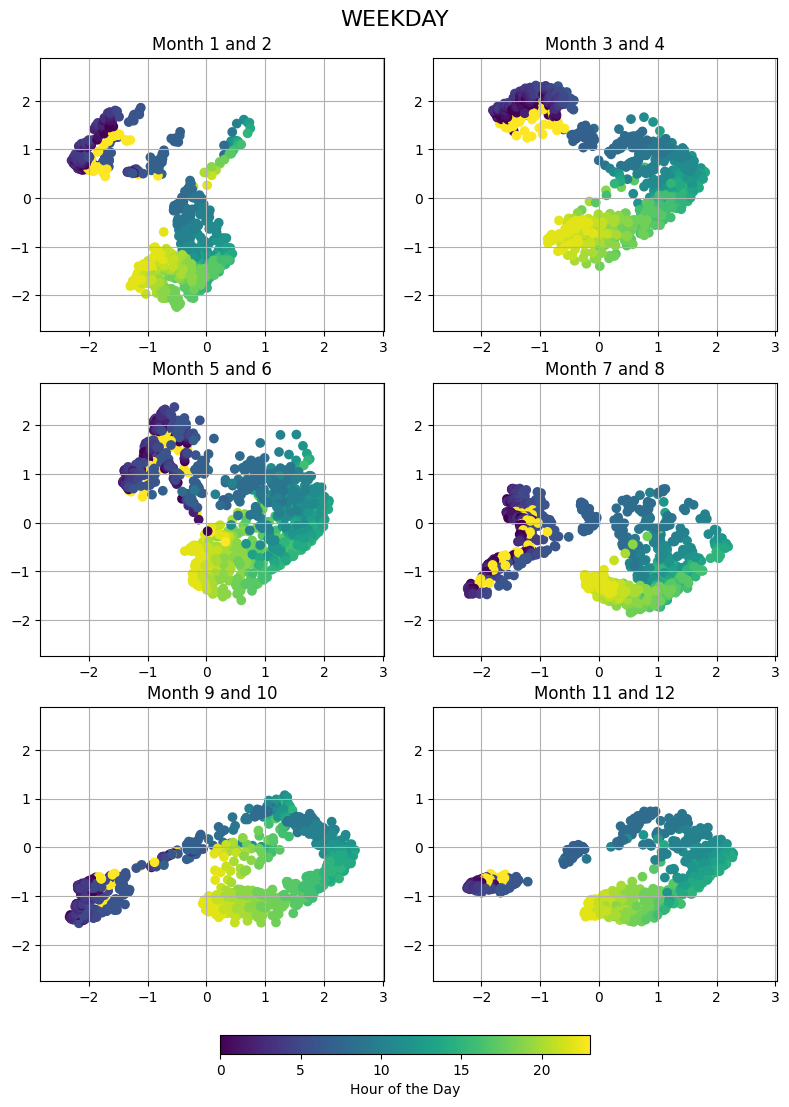}
    \caption{Weekday}
  \end{subfigure}
  \hfill
  \begin{subfigure}[t]{0.49\textwidth}
    \centering
    \includegraphics[width=\textwidth]{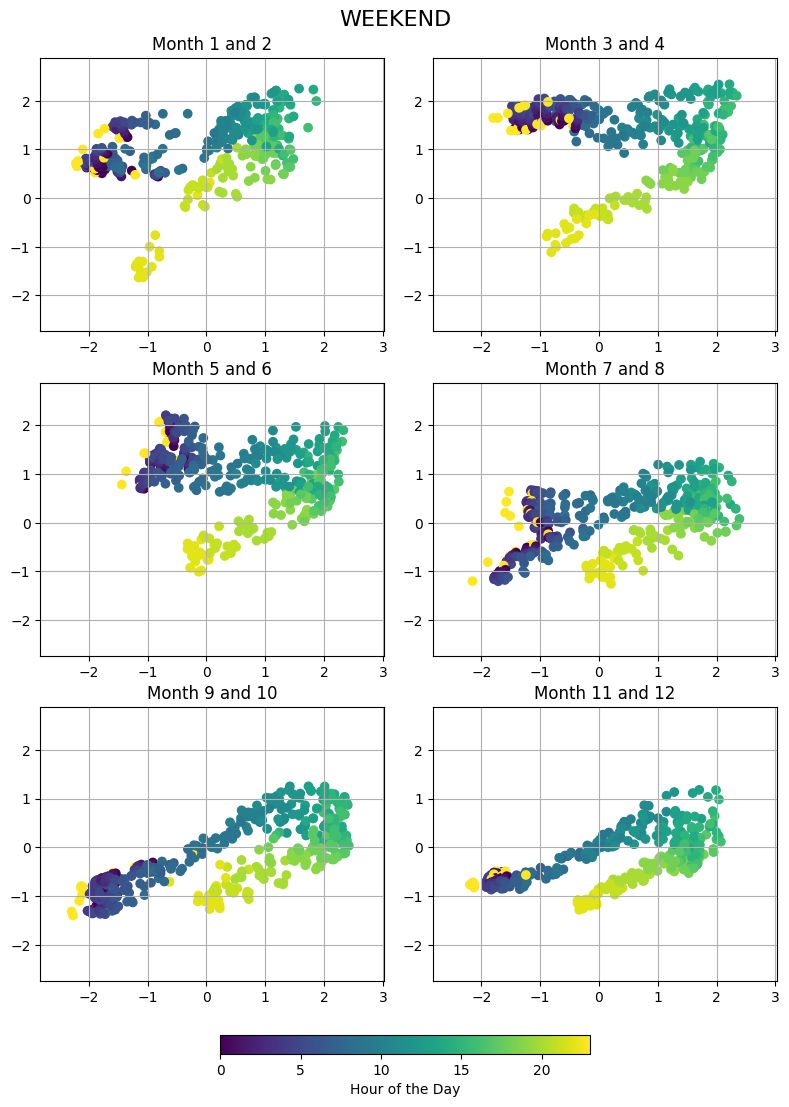}
    \caption{Weekend}
  \end{subfigure}
  \caption{Comparison of Urban Vibrancy Embedding on Weekdays and Weekends of year 2017.}
  \label{fig:urban-vibrancy-monthly}
\end{figure*}

\section{Result}

\subsection{Interpretation of Urban Vibrancy Embedding}

In this paper, real-time LTE floating population data, which reflects urban vibrancy, is spatio-temporal data. To facilitate a deeper understanding of this data and to enable multi-faceted interpretation, we employed data visualization techniques. Principal Component Analysis (PCA) was used to interpret and analyze the data through visualization.

\subsubsection{Embedding Analysis of Temporal Urban Vibrancy Patterns}

Figure~\ref{fig:urban-vibrancy-pca-component} reveals distinct embedding patterns for weekdays and weekends, displayed in subplots (a) and (b), respectively. This facilitates a clear comparison between these temporal contexts. Figures~\ref{fig:pca-component-1} and \ref{fig:pca-component-2} showcase the decoded results of PCA1 and PCA2. Plus, we find 5 PCA components can represent the most influential components in the dataset based on reconstruction error with a 99.7\% threshold.

In South Korea, urban areas are categorized into residential, commercial, industrial, and green zones, while non-urban areas include managed, agricultural/forestry, and natural conservation zones. PCA1 primarily reflects population activity in Seoul's commercial and industrial zones, whereas PCA2 captures activity in residential, green, and non-urban zones. Thus, PCA1 and PCA2 serve as proxies for population movement within these distinct urban and non-urban sectors, highlighting temporal urban vibrancy patterns that vary by zone type. 

\subsubsection{Temporal Evolution and Seasonal Variations}

Figure~\ref{fig:urban-vibrancy-monthly} shows the temporal evolution of urban vibrancy embeddings for weekdays and weekends, visualized at bimonthly intervals. Weekday embeddings tend to form concentrated clusters, indicating population convergence in specific areas due to work and commuting routines. Conversely, weekend embeddings are more dispersed, suggesting a broader array of activities with less structured movement as people engage in leisure and recreational pursuits.

A color gradient from purple (00:00) to yellow (23:00) illustrates the daily temporal flow, revealing a rotation in embedding values over time. Peak activity is observed from 10:00 to 15:00, while reduced activity occurs from 00:00 to 05:00, indicating population migration outside Seoul. Notably, from 20:00 to 23:00, a sharp increase in activity reflects significant movement in and out of Seoul. Weekday patterns show counterclockwise rotations, while weekend patterns display clockwise rotations with a noticeable upward trend from January to June in areas like parks and the Han River, due to outdoor activities.

Seasonal variations are also evident; winter months (January and February) show more concentrated clusters, while summer months (July and August) exhibit broader distributions, reflecting increased outdoor activities during warmer periods. These seasonal shifts correspond to changes in mobility, with summer promoting more outdoor movement and dispersed activities, while colder winter months limit outdoor engagements.

\subsubsection{Urban Planning Implications and Impact of Unique Events}

The observed temporal patterns provide valuable insights for urban planning and management. Weekday clustering suggests that urban transportation and infrastructure can be optimized to meet predictable commuting demands. In contrast, the dispersed weekend patterns call for a more flexible approach to accommodate varied leisure activities. Additionally, businesses might consider adjusting their hours or staffing levels to align with these urban vibrancy trends, particularly during weekends with heightened activity.

Unique events can also significantly affect urban vibrancy. For instance, from January to March 2017, large-scale protests of impeachment of ex-president in Seoul drew approximately 800,000 people to Jongno-gu. The embeddings from this period show unique rotational patterns compared to 2018 data, reflecting the impact of this rare collective event on urban dynamics. These insights emphasize the need for responsive planning strategies that account for both routine and exceptional events, supporting a more adaptive and resilient urban environment.

In summary, this study offers a detailed view of urban vibrancy, capturing how movement patterns fluctuate across weekdays, weekends, and seasons. The findings provide valuable guidance for urban planning, infrastructure management, and public policy, contributing to a more adaptable urban environment that meets the diverse needs of city inhabitants.

\input{tables/tab-performance}

\subsection{Traffic Prediction Enhancement}

Table~\ref{tab:performance} compares the performance of various models (RNN, DCRNN, GTS, GMAN) with different embeddings (None, TE, VE, TVE) for predicting SUBWAY, VOLUME, and SPEED metrics over horizons of 1, 2, 3, and 6. 

The RNN model shows high errors without embeddings, with metrics like SUBWAY errors at 352.47 (horizon 1) and SPEED errors at 6.133. Adding TE, VE, or TVE improves performance, with TVE generally yielding the best results, such as reducing SUBWAY error to 73.34. DCRNN performs better than RNN, especially with VE, achieving a SUBWAY error of 91.26 (horizon 1) and SPEED error of 4.38. GTS generally has lower errors and benefits significantly from VE and TVE, with SUBWAY errors dropping to 57.77 and VOLUME errors to 138.21 (horizon 6) with TVE. GMAN shows good performance for VOLUME (169.41 with VE) and SPEED predictions but has higher SUBWAY errors compared to GTS and DCRNN.

Among embeddings, TE alone reduces errors but is outperformed by VE and TVE. VE improves predictions across all models, especially with DCRNN and GTS. TVE achieves the lowest errors in most cases, like a 57.77 SUBWAY error for GTS, combining temporal and vibrancy data effectively.

Errors generally increase with horizon length, but embeddings like VE and TVE help control this, particularly for VOLUME and SPEED. For specific metrics, GTS with VE or TVE performs best for SUBWAY, while DCRNN with VE or TVE excels in SPEED predictions.

In summary, embedding strategies, especially VE and TVE, significantly enhance predictive accuracy, particularly for GTS and DCRNN models. VE alone is effective, but TVE provides the lowest errors by adding a comprehensive context.

% Table~\ref{tab:performance}는 Urban Vibrancy Embeddsing(VE)을 넣었을 때의 실험 결과를 나타낸다. 

% VE를  적용했을 때, Traffic Volume, Subway Demand, Traffic Speed 순으로 교통 예측 결과가 향상됨을 알 수 있으며, 대부분의 경우에서 VE를 포함했을 때 성능이 향상되었다.

% %GMAN 재실험

% Traffic Volume 의 경우 DCRNN 을 제외한 모든 모델에서, VE가 적용되었을 때 outstanding한  성능을 보이는 것을 알 수 있음. 
% 1,2,3 시간 이후의 예측에서 모두 오차율을 개선하였음. 
% 특히 6시간 뒤 예측에서 오차율을 NONE 보다 GRU 13.8\%, DCRNN 19.6\%, GTS 32.2\%, GMAN 63\%, STTN 16.2\% 개선했음. 
% 6시간 뒤 예측에서 DCRNN은 STE가 적용되었을 때 가장 높은 성능을 보였지만, TVE 적용보다 0.5\% 높은 오차율로, 근소한 차이를 보였음.
% GRU, GTS, STTN의 경우 TE 및 SE 를 모두 포함했을 때, 오차율이 가장 낮게 나타났으며, GMAN의 경우는 TE를 제외하고 SVE가 적용되었을 때 오차율이 가장 낮게 나타났다.

% ---
% Subway 의 경우 GTS, GRU에서 STUE, TUE 일때 가장 높은 성능을 보임.
%  STTN을 제외한 나머지 모든 모델에서, VE가 적용될 경우의 성능 향상이 관찰되었음. 특히 DCRNN의 경우 6시간 뒤 예측에서 SVE의 오차율이 TE가 적용된 경우보다 66.5\% 향상되었음. 

% ----

% Traffic Speed 의 경우 GMAN을 제외하고 VE를 적용했을 때, STE, TE에 비해 오차율이 개선되지 않음을 볼 수 있음. GMAN 의 경우 2, 3, 6 시간 뒤 예측에서 가장 낮은 오차율이 나타났음. 전체적으로 STE 및 TE가 적용되었을 때 오차율이 개선되는 것이 나타났으나, VE 및 SVE, STVE 를 적용했을때 오차율이 크게 개선되지 않는 것을 볼 수 있었음. 따라서 speed 예측의 VE 적용은 STE, TE에 관한 데이터가 손실되었거나, 이를 적용할 수 없는 경우에 고려될 수 있음을 알 수 있음.

% \subsection{Case Study}

% VAE reconstruction loss를 줄이기 위해 어떻게 했는지를 discussion 한다. (Ablation Test)

%% file: tables/tab-performance.tex
\begin{table*}[t]
\caption{Performance Comparison} \label{tab:performance}
\centering
\begin{tabular}{@{}ll|rrrr|rrrr|rrrr@{}}
\toprule
      &         & \multicolumn{4}{c|}{SUBWAY}                                            & \multicolumn{4}{c|}{VOLUME}                                            & \multicolumn{4}{c}{SPEED}                                             \\ \midrule
      & Horizon & 1               & 2               & 3               & 6               & 1               & 2               & 3               & 6               & 1               & 2               & 3               & 6               \\ \midrule
RNN   & NONE    & 352.47          & 352.96          & 351.77          & 351.71          & 196.39          & 212.95          & 220.95          & 231.52          & 6.1336          & 6.6175          & 6.9996          & 7.2999          \\
      & TE      & 77.33           & 80.59           & \textbf{82.31}  & \textbf{84.11}  & 190.47          & 197.21          & 201.07          & 209.95          & \textbf{5.7842} & \textbf{5.9992} & \textbf{6.1433} & \textbf{6.3660} \\
      & VE      & 78.76           & 85.26           & 92.26           & 98.22           & 189.87          & 197.06          & 200.50          & 204.83          & 6.1837          & 6.9754          & 7.5057          & 7.7835          \\
      & TVE     & \textbf{73.34}  & \textbf{77.44}  & 83.29           & 90.54           & \textbf{187.40} & \textbf{193.68} & \textbf{196.81} & \textbf{199.58} & 5.7868          & 6.0699          & 6.4011          & 6.8643          \\ \midrule
DCRNN & NONE    & 142.20          & 207.63          & 260.95          & 359.25          & 152.84          & 187.81          & 206.16          & 239.85          & 5.3059          & 6.7597          & 7.4098          & 8.4015          \\
      & STE     & 131.19          & 190.28          & 234.26          & 288.31          & \textbf{145.08} & \textbf{164.11} & \textbf{174.69} & \textbf{191.94} & \textbf{4.3802} & \textbf{5.1023} & \textbf{5.4427} & \textbf{5.9109} \\
      & SVE     & \textbf{91.26}  & \textbf{102.45} & \textbf{111.70} & \textbf{120.22} & 163.80          & 183.82          & 192.08          & 202.13          & 4.7570          & 5.5517          & 5.8617          & 6.2579          \\
      & STVE    & 192.80          & 234.67          & 258.78          & 284.82          & 162.63          & 176.39          & 183.17          & 192.94          & 4.4395          & 5.2042          & 5.5707          & 5.9860          \\ \midrule
GTS   & NONE    & 69.68           & 79.80           & 85.14           & 91.10           & 161.11          & 210.87          & 235.88          & 276.99          & 4.7113          & 5.6962          & 6.1930          & 6.7567          \\
      & STE     & 65.98           & 70.54           & 73.76           & \textbf{79.12}  & 142.23          & 166.38          & 175.12          & 190.28          & \textbf{4.2430} & \textbf{4.9347} & \textbf{5.2724} & \textbf{5.7986} \\
      & SVE     & 62.73           & 72.29           & 77.92           & 83.53           & 155.11          & 179.22          & 188.87          & 200.35          & 4.4471          & 5.2555          & 5.6333          & 6.0569          \\
      & STVE    & \textbf{57.77}  & \textbf{67.32}  & \textbf{72.87}  & 80.38           & \textbf{138.21} & \textbf{163.32} & \textbf{174.82} & \textbf{187.72} & 4.3116          & 5.0549          & 5.4554          & 5.9742          \\ \midrule
GMAN  & NONE    & 523.22          & 515.00          & 512.95          & 533.61          & 426.75          & 341.96          & 311.69          & 549.61          & 10.0531         & 9.3465          & 9.2520          & 13.0090         \\
      & STE     & 312.69          & \textbf{307.75}          & \textbf{300.90}          & \textbf{277.29}          & 169.41          & 184.02          & 192.87          & 205.50          & 5.1138          & 5.6817          & 5.9832          & 6.4349          \\
      & SVE     & 566.58          & 567.34          & 565.11          & 564.72          & \textbf{164.71} & \textbf{177.08} & \textbf{185.95} & \textbf{200.01} & \textbf{5.0487} & 5.6810          & 6.0083          & 6.6369          \\
      & STVE    & \textbf{307.84} & 312.39 & 309.23 & 311.89 & 165.76          & 178.50          & 186.91          & 199.11          & 5.0951          & \textbf{5.6375} & \textbf{5.9554} & \textbf{6.3519} \\ \bottomrule
\end{tabular}
\end{table*}

%% file: sec7-conclusion.tex
% \section{Discussion}

% Multi-channel Embedding

\section{Conclusion}

In conclusion, this research presents a novel framework for enhancing traffic prediction through the use of Urban Vibrancy Embeddings derived from real-time floating population data. By employing Variational Autoencoders (VAE) to compress high-dimensional urban data into meaningful embeddings and integrating these with Long Short-Term Memory (LSTM) networks within a sequence-to-sequence forecasting framework, we capture and leverage the dynamic patterns of urban vibrancy effectively.

Our findings highlight the value of this approach in several key areas. First, the use of principal component analysis (PCA) allows for an intuitive interpretation of the temporal patterns embedded in urban vibrancy data, such as differences between weekdays and weekends, as well as seasonal and hourly shifts. Second, the dynamic embedding of real-time urban data improves model responsiveness and adaptability to fluctuating urban conditions, which is crucial for applications in smart city environments. Third, our experiments demonstrate notable improvements in the predictive accuracy of models like RNN, DCRNN, GTS, and GMAN, showing the versatility and effectiveness of our approach across different modeling architectures.

Through rigorous validation on real-world data from Seoul, we confirm the practical applicability and robustness of our method. By releasing our code and dataset, we aim to support reproducibility and encourage further exploration in this area. This study advances the field by providing a more nuanced and responsive approach to traffic prediction, integrating urban vibrancy insights to enhance the adaptability of models to the dynamic realities of urban mobility.